\documentclass[sigconf]{acmart}

\usepackage{booktabs} 

\usepackage{balance}

\renewcommand\footnotetextcopyrightpermission[1]{} 
\begin{document}

\title{Automation of Feature Engineering for IoT Analytics}

\author{\textbf{Snehasis Banerjee}}
\affiliation{%
  \institution{TCS Research \& Innovation}
  \streetaddress{Tata Consultancy Services}
  \city{Kolkata} 
  \state{West Bengal} 
  \postcode{700160}
}
\email{snehasis.banerjee@tcs.com}

\author{\textbf{Tanushyam Chattopadhyay}}
\affiliation{%
  \institution{TCS Research \& Innovation}
  \streetaddress{Tata Consultancy Services}
  \city{Kolkata} 
  \state{West Bengal} 
  \postcode{700160}
}
\email{t.chattopadhyay@tcs.com}

\author{\textbf{Arpan Pal}}
\affiliation{%
  \institution{TCS Research \& Innovation}
  \streetaddress{Tata Consultancy Services}
  \city{Kolkata} 
  \state{West Bengal} 
  \postcode{700160}
}
\email{arpan.pal@tcs.com}

\author{\textbf{Utpal Garain}}
\affiliation{%
  \institution{Indian Statistical Institute}
  \streetaddress{203, Barrackpore Trunk Road}
  \city{Kolkata} 
  \state{West Bengal} 
  \postcode{700108}
}
\email{utpal@isical.ac.in}

\begin{abstract}
This paper presents an approach for automation of interpretable feature selection for Internet Of Things Analytics (IoTA) using machine learning (ML) techniques. Authors have conducted a survey over different people involved in different IoTA based application development tasks. The survey reveals that feature selection is the most time consuming and niche skill demanding part of the entire workflow. This paper shows how feature selection is successfully automated without sacrificing the decision making accuracy and thereby reducing the project completion time and cost of hiring expensive resources. Several pattern recognition principles and state of art (SoA) ML techniques are followed to design the overall approach for the proposed automation. Three data sets are considered to establish the proof-of-concept. Experimental results show that the proposed automation is able to reduce the time for feature selection to $2$ days instead of $4-6$ months which would have been required in absence of the automation. This reduction in time is achieved without any sacrifice in the accuracy of the decision making process. Proposed method is also compared against Multi Layer Perceptron (MLP) model as most of the state of the art works on IoTA uses MLP based Deep Learning. Moreover the feature selection method is compared against SoA feature reduction technique namely Principal Component Analysis (PCA) and its variants. The results obtained show that the proposed method is effective.
\end{abstract}



\maketitle

\vspace{1in}

\textit{Copyright retained by the authors.}

\section{Introduction}
Internet of Things analytics (IoTA) involves a lot of applications  \cite{2014cikmSnehasis} \cite{2013wam} \cite{2012www} \cite{2013WIsnehasis}   in health, wellness, sustainability, transportation, smart city administration, and urban health while deploying any industrial sensor systems. Some of such applications in health care like blood pressure classification, Coronary Artery Disease (CAD) classification or machine health classification (into good condition or bad condition) involves classification as the machine learning (ML) task \cite{Bengio:Book}. It is required to identify some characters, known as feature, of the input signal in order to classify it using popular ML methods like Support Vector Machine (SVM) or Random Forest (RF). But the method of identifying a set of suitable features for the classifier is a time and cost absorbing task as it needs mostly domain expertise of the signal processing and IoT expert. Hiring people with such niche skill set is costly. Another problem is that some times the features listed  by the domain expert are not relevant and if the data set includes less number of instances with a lot of features, the curse of dimensionality comes into play. This results into the need for feature reduction which can be accomplished into two different ways namely feature dimension reduction \cite{Ullman:book} or reducing the number of features by feature selection \cite{Maji:paperMRMSexp}, \cite{Peng:MRMR}. These two steps of listing features and reducing the number of features is known as Feature Engineering (FE) as a whole. Another approach to get rid of this issue of FE is to use some Deep Learning (DL) techniques like  Multi Layer Perceptron (MLP) with fully connected layers or Convolutional Neural Network (CNN) which can automatically identify suitable feature set for the classification. But DL methods makes an assumption of having a huge number of annotated data-set to train the system which is mostly not available for IoT tasks. DL remains as a very popular technique for computer vision and natural language processing (NLP) and speech recognition. Another popular method is using Principal Component Analysis (PCA) to derive features that have high correlation to the class. However, in both cases interpretation of the features is an issue, and the domain expert cannot take advantage to carry out causal analysis for a given problem, once suitable features are identified. 
For example, a cardiologist can justify that the PPG (photoplethysmogram) signals should contain a signal at a particular frequency range to classify a CAD signal. But neither DL methods nor the feature dimension reduction techniques can interpret the features as the feature space do not directly relate to physical world. So in this paper authors have proposed a method of FE for classification using ML techniques where a Feature Listing Database is maintained that becomes useful once good features are identified. The contributions of the proposed work is enlisted:\\
1. A survey has been carried out to understand the workflow of a typical IoT System development and identify the stages that demand maximum effort and niche skills, thereby costing more money and time.\\
2. The work proposes a method for automation of FE so that the suitable features for classification can be obtained in less time, less cost and the recommended features are physically interpretable to assist the domain expert in making casual analysis.\\
3. A WIDE architecture for feature listing is proposed that starts to use basic features in the basic layer and extracts more derived features on the higher layer. This layered architecture also helps to reduce the complexity as the features for different layers are computed iteratively. Features of higher layer are not computed once the desired performance is obtained at a lower layer. The performance is measured using different metrics like sensitivity, specificity, F-score.\\

The proposed method is tested against annotated data sets on man and machine predictive analytics, the use cases of which demand interpretation of features. The used data sets contains (i) NASA's bearing data set that includes good and bad bearing data, (ii) Emotion dataset to classify emotion of users, and (iii) MIMIC-II data set to classify high and low blood pressure from PPG signal. The accuracy obtained by applying some classifier on the recommended feature set are compared against the performance obtained by applying the same classifier on the SoA features reported in literature to solve these point problems. Some of these point problems were also tried in our research lab and hence we were able to compare the development time. We have also compared the proposed method against DL techniques like many layered Multi Layer Perceptron (MLP) method as most of the literatures have used MLP for sensor signal processing \cite{voyant:mlpTime}. The proposed feature selection technique is compared against a popular feature dimension reduction technique namely Principal Component Analysis (PCA). Finally, through an example we have also given a physical interpretation of the recommended features so that the recommended features can be validated against the physical phenomenon.

\begin{figure}[t]
\centering{
\includegraphics[width=5cm]{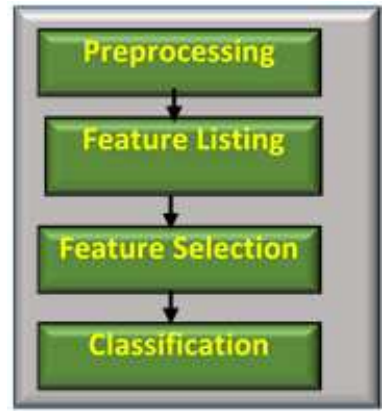}
\caption{High level workflow for IoT Analytics}}
\label{fig:workflow}
\end{figure}
\begin{figure}[!htb]
\centering{
\includegraphics[width=9cm]{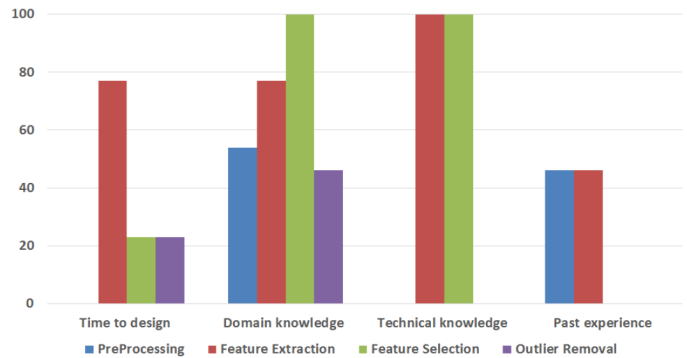}
\caption{Analysis of pain areas in IoT Analytics}}
\label{fig:analysis}
\vspace{-1.5em}
\end{figure}

Section 2 discusses about the survey conducted to analyse typical IoTA tasks. Section 3 describes the process and corresponding system architecture. Section 4 discusses the various experiments carried out. Finally section 5 concludes the paper.

\section{Background}

\begin{table*}[!t]
\small
\centering
\begin{tabular}{|l|c|c|c|}
\hline

Sensor &Signal Type &Goal &Steps \\
\hline
Camera, mobile, Kinect, &Stochastic  &Object and Scene &Sampling$\rightarrow$Calibration$\rightarrow$Object  \\
 &  & Recognition & Classification \\
\hline
Hyper spectral sensors, &NA  &Object recognition &Image acquisition$\rightarrow$Pre-Processing $\rightarrow$\\
 &  & &Calibration$\rightarrow$ Feature listing$\rightarrow$\\
 &  & &Supervised or Unsupervised learning \\
\hline
Biosensors like ECG, PPG, &NA  &Object recognition &Pre-Processing $\rightarrow$Feature Listing\\
EEG, EMG, Camera, &  & &$\rightarrow$Classification \\
\hline
Camera &NA &Object Recognition &Pre-Processing $\rightarrow$Feature Listing\\
&  & &$\rightarrow$Classification \\
\hline
Microphone, Camera, &Periodic, Stationary,  &Physiological &Pre-processing$\rightarrow$Signal Quality Checker \\
IR Camera  &Non-stationary &condition estimation  &$\rightarrow$Feature Listing$\rightarrow$Feature \\
           &  & &Selection$\rightarrow$Computation   \\
\hline
Thermal camera,  &Multiple frequencies &Classification, Estimation &Filtering$\rightarrow$ Denoising $\rightarrow$ \\
LED/PhotoDiode & &calculation, imaging &features detection $\rightarrow$classification \\
& &beamforming, denoising & \\
\hline
EEG, GSR, SpO2, &Aperiodic, Non-stationary & Classification and &Pre-processing$\rightarrow$Noisy window removal  \\
Camera, & &clustering, estimation &$\rightarrow$Feature Listing$\rightarrow$Feature Selection \\
Eye tracker & & &$\rightarrow$Computation \\
\hline
\end{tabular}
\caption{Steps involved in IoT sensor signal processing and analytics for different applications}
\label{tab:steps}
\vspace{-1.5em}
\end{table*}

We made a survey over the associates within our organization involved in the IoTA to identify the pain areas of the application developers. In this survey, we have surveyed seven ($7$) projects which include ninety-five ($95$) employees (female: $35$ and male: $60$). The associates under each project form a team. Each team consists of members among whom 30\% having more than 10 years of experience and Master's or higher academic qualification and each team is mostly led by a Ph.D. person. Sixty per cent ($60$\%) associates in each team have 5-10 years of experience and expertise in signal processing and the remaining $10$\% are developers mostly equipped with good coding skill in C/Java/Python but don't have in depth signal processing, IoT or domain knowledge.

Seven projects involved in this survey are (i) Motion capture, object recognition and rendering of articulated objects from Kinect based skeleton data, (ii) Object classification and recognition from Hyper-spectral sensor, multi-spectral sensors (Landsat, digital globe WorldView etc.)  (iii) Bio sensors based object recognition, (iv) Outdoor camera based object recognition. (v) Detecting heart rate and blood pressure from PPG,  (vi) Thermal imaging based object classification, and (vii) EEG (electroencephalogram) based cognitive load classification. For each project, the project goal, sensors used, and the steps followed to achieve the goal are shown in Table \ref{tab:steps}. These steps can be combined to construct a superset of steps followed in any sensor signal processing based IoTA as shown in Figure 1. The steps used in this figure are briefly described here:
\begin{itemize}
\item{Pre-Processing}: This module aims to filter out the noises in a signal. It also helps to remove the outlier in the signals. Different pre-processing techniques are available in the literature.
\item{Feature Listing}: Any signal is represented by some features. For example, time domain or frequency domain analysis normally give many features for an input signal.
\item{Feature Reduction}: The possible number of features for a sensor signal is very huge and so it is required to reduce the number of features. Feature reduction techniques should be selected with the following points kept in mind: (i) Interaction among the features, (ii) Interaction among the features with machine learning tool, (iii)	Goal to achieve which can be obtained from an annotated data set, (iv)	Interpretability of the reduced feature.
\item{Classification}: The technique takes the experience which is a representation of the feature set and the label as input and optimizes some parameters like accuracy, sensitivity to accomplish a goal like classification.
\end{itemize}

The analysis as shown in Figure 2 clearly reveals that most of the associates under survey express that feature listing and feature selection or feature reduction requires the maximum domain knowledge and technical knowledge. The other steps like pre-processing (including noise cleaning and outlier removal) takes the highest time to design. As the pre-processing algorithms are well established and even source code for them are available in the web, we propose to provide the developer with a list of source codes for all those different state of the art preprocessing techniques so that the developer can try any of the methods on their data set and select the best one among them. One such method is described in \cite{TCS:Sensys}. But feature selection is a difficult process. The input to our proposed automated system is the labeled data set of the problem to solve. Another aspect of the study was to list down the features, commonly reported in the related literature, that are used in IoTA applications and finally recommend the features to be used for the given use case.\\
\section{Proposed Method}

\begin{figure*}[ht]
\centering
\includegraphics[width=6.0in]{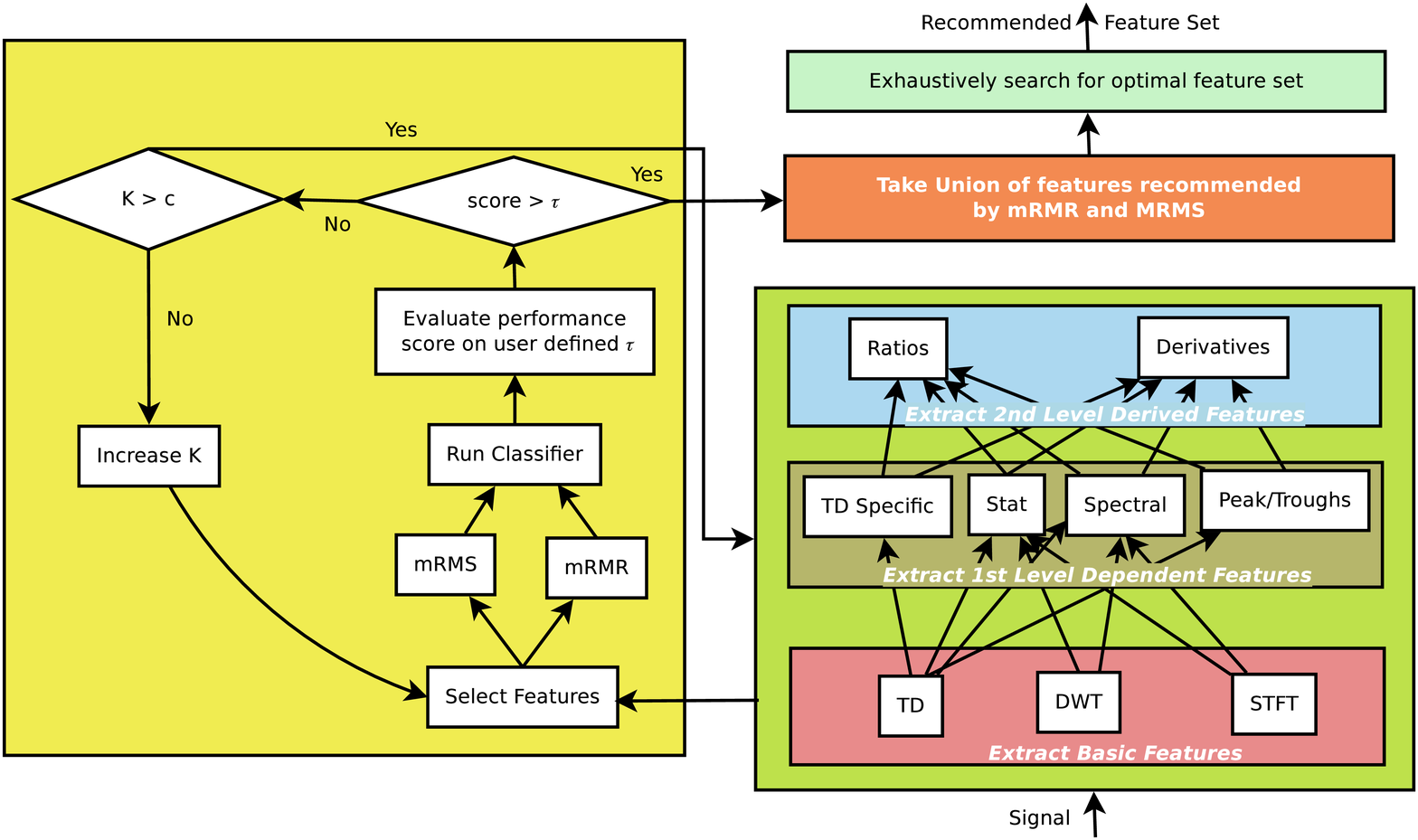}
\caption{Proposed Method of Feature Recommendation}
\label{fig:FeatureEngineering}
\end{figure*}

In this section, we are going to discuss how to automate the feature selection step which is the most time consuming and domain specific expertise dependent step. We plan to use state of the art pattern recognition techniques to achieve this. The key idea behind this is as follows.

{\it Exploration of all possible features}: Finding the right set of features for any pattern classification task is still an unsolved problem. Great amount of research is still concentrating on this problem \cite{TCS:Sensys}. In our approach, we, at first, have studied the existing literature~\cite{Lu:3} - \cite{Coutrix:20} to have a more or less exhaustive list of features which have been used by different researchers for different classification tasks on different sensor data sets.

We have organized these reported features in an hierarchical manner as shown in Figure ~\ref{fig:FeatureEngineering}. The features reported in literature can be mainly classified in three types: (i) time domain features (TD) (ii) Fourier transformation based features (FD) (iii) Discrete Wavelet transformation based features (DWT). A challenge of using features from Wavelet Transform is the appropriate selection of a suitable mother wavelet. This is because more than 100 different types of mother wavelets are reported in literature, and a ready made automatic wavelet selection method or tool do not exist. Hence, considering the the different feature types, it is possible to get a large number (say, $N$) of features (including coefficients of the transform domain) from the sensor signals. This results in $2^N-1$ possible combinations of features.

{\it Feature Selection}:
The value of $N$ could be as high as $6,000,000$ as explained later in this section.  Therefore, in order to find the optimal feature set, exploitation of all such $2^N-1$ combinations is practically infeasible as this would require testing of machine performance for each combination. This can be achieved mostly by the domain experts from their experience and understanding about the physical phenomena. Typically feature selection method takes 4-6 months for such problems. In pattern recognition literature, there are several feature reduction techniques available which can give a reduced set of features giving optimum performance \cite{Chattopadhyay:ICIP}, \cite{2013tanuICDAR}, \cite{2011tanuICCE}. Principle Component Analysis (PCA) is one such commonly used technique for feature reduction. Actually, what PCA does is feature extraction, i.e. the resultant features are not interpretable. However, we need interpretable features for our tasks. For instance, in health analytics, if we know that certain features are very strong in predicting CAD (coronary artery disease) patients, the doctors need to know what these features are. Therefore, in order to reduce the number of feature to make the feature recommendation task tractable we follow feature selection instead of feature extraction as feature selection methods do not alter the original features.

In our method, we followed an iterative feature selection where $k$-features are selected at each iteration and system performance (e.g. classification accuracy) is checked for this feature set. If the selected feature set results in {\it expected} performance, that feature set is recommended and process halts. Else, another set of $k$-features is selected in the succeeding iteration and similar steps are carried out. For checking the classification accuracy, we choose SVM-(support vector machines) based classification with different kernels. Different values of $k$ are fitted in progressive increments to get a good result. For a given value of $k$, features are selected using two techniques namely, MRMS \cite{Maji:paperMRMS} and mRMR \cite{Peng:MRMR}. The reason for choosing these two techniques is their impressive efficiency in feature selection as demonstrated in~\cite{Maji:paperMRMSexp}. Details on mRMR and MRMS used for this framework can be found in \cite{2016nipsSnehasis}.

\subsection{The proposed Architecture for Feature Recommendation}
The proposed method is applicable in extracting and recommending features from 1D sensor signals. The sensor signal is subjected to time domain, frequency domain and time frequency domain analysis to extract features which for example counts to almost 12 million features for input data of size 20,000. Let us assume a 1D signal of length n sampled at fs frequency. So after subtracting the mean from the signal we get the same n number of TD features. Now the entire signal is splitted into multiple overlapping windows to extract the FD features. So if we consider a 1sec window with 50\% overlap then we get (n*2)/fs number of windows. Considering the STFT window size to be 256, number of FD features can be obtained is (256*n*2)/fs. Similarly by applying DWT we obtain another n features. So extracting first level features we get n+n+(512*n)/fs $\approx$ 3n number of features. The level two features are derived from each window of the level one features. Thus in level 2 around 20*(3n/fs)*2$\approx$120n features are extracted. Therefore around 3n+120n=123n features are extracted at the end of layer2. Similarly in 3rd layer features are derived from each window of 2nd level features which is around 2*(120n/fs)*2 $\approx$ 480n. So after 3rd layer the number of features extracted is of the order 123n+480n $\approx$ 600n. Now in dataset 1, value of n is nearly 20,000. So number of features extracted after 3rd layer is nearly 600X20,000$\approx$ 12,000,000.
Level two features are applied on top of each of the level one features. Thus after level 2 this number of features extracted is becoming nearly of the order of 4n*4=16*n. Similarly after 3rd layer this number becomes of the order or 30*n. Now in data set 1 this n is nearly 20,000. So after 3rd layer of feature extraction total number of feature extracted from the signal is nearly $30\times20,000 \approx$ 6,00,000.
So, it is not possible to apply any exhaustive search on top of this 12,000,000 features so that one can optimize the performance metrics say sensitivity or specificity. The feature selection architecture used in this paper is shown in Figure~\ref{fig:FeatureEngineering}. The proposed method is as described below: \\
Input: Time domain signal \\
Output: Recommended feature list
\begin{itemize}
\item{} Compute the mean subtracted time domain signal by subtracting the mean from the original signal. This entire signal is considered as the time domain (TD) feature.
\item{} Compute the short term Fourier transform(STFT) for each window where window size is provided as an user input.
\item{} The given signal is subjected to four level Discrete Wavelet Transform (DWT) using the most optimal mother wavelet. The optimal mother wavelet is selected using the property of maximum energy to entropy ratio  \cite{ngui2013wavelet}.
\item{} Take the union of the all three TD, FD, and DWT features. Let this set be defined as $FL_{1}$
\item{} Apply minimum redundancy maximum relevance (mRMR) and Maximum Relevance Maximum Significance (MRMS) for $FL_{1}$. Each of these methods recommends a different feature set say x and y.
\item{} Iteratively select the number of features to be recommended by each of these methods so that the recommended features applied on a classifier can exceed the performance metric score defined by the user, say $\tau$.
\item{}  Take union of the features recommended by both mRMR and MRMS methods. Let the recommended feature set be z where $z = x \cup y$.
\item{} Features from level 2 ($FL_{2}$) and level 3 ($FL_{3}$) are extracted using the algorithms shown in Figure \ref{fig:FeatureEngineering}.
\item{} In level 2 we have extracted spectral features, statistical features, and peak-trough features. Spectral features used in the proposed method are Spectral Centroid, Spectral Crest Factor, Spectral Decrease, Spectral Flatness, Spectral Flux, Spectral Kurtosis, Spectral Rolloff, Spectral Skewness, Spectral Slope and Spectral Spread which are computed for each window of mean subtracted signal. Statistical Features used here are mean, variance, standard deviation, root mean square, skewness, kurtosis. Average peak amplitude, average trough amplitude, average peak to peak distance and average trough to trough distance are computed as the peak-trough features. $FL_{3}$ includes different ratios and derivatives of the $FL_{2}$ features.
\item{} These $FL_{2}$ and $FL_{3}$ features are also reduced using mRMR and MRMS similarly.
\item{} Once the feature space is reduced, exhaustively generate all possible combinations of features and apply classifier with different parameters.
\item{} Apply Support Vector Machines (SVM) classifiers with different kernels namely (i) linear, (ii) radial basis function (RBF), (iii) sigmoid, and (iv) polynomial and each of the kernels is tested with different parameter values.
\item{} Thus construct the model for classification by selecting the feature list and the SVM kernel for which the recognition accuracy maximizes. SVM is selected as a classifier as the problems at hand are binary classification where SVM is known to excel and converge quickly.
\end{itemize}

\section{Experiments}

\begin{table*}
\caption{Comparison of proposed method against state of the art for different data sets}
\label{Table:comparison}
\centering
\begin{tabular}{|l|c|c|c|c|c|c|}
\hline
Data set &SOA no of  &Recommended no  &SOA  &Recommended  & SOA  &Recommended \\
         &features   &of features     &accuracy &accuracy &effort &effort \\
\hline
NASA data set &15 &10 &99.38 &100 &6 months &2 days \\
\hline
Emotion data set &16 &11 &82.3 &90.91 &4 months &2 days \\
\hline
BP data set &23 &15 &79.5 &87.8 &6 months &2 days \\
\hline

\end{tabular}
\end{table*}
\begin{table*}
\caption{Comparison of proposed method against 5-layered MLP}
\label{Table:comparison}
\centering
\begin{tabular}{|l|c|c|c|c|c|c|}
\hline
Data set &Activation  &No. of Epoch &MLP Accuracy  &SoA Accuracy &Proposed\\
         &Function    &				&		 	  &		 & Method Accuracy 		 \\
\hline
NASA data set &softmax   &5         &   71.00             &99.38 &100  \\
\hline
Emotion data set &relu   &15         &   50.0             &82.3 &90.91  \\
\hline
BP data set &softmax   &10         &   50.0             &79.5 &87.8  \\
\hline

\end{tabular}
\end{table*}
\begin{table*}
\caption{Comparison of proposed method against PCA}
\label{Table:comparison}
\centering
\begin{tabular}{|l|c|c|c|c|c|c|}
\hline
Data set &Dimensionality  		   &No. of Principal &PCA Accuracy  &SoA Accuracy &Proposed\\
         &Reduction Algorithm  &Components				&		 	  &		 & Method Accuracy 		 \\
\hline
NASA data set &svd   &5         &   94.00          &99.38 &100  \\
\hline
Emotion data set &svd   &10        &   50.0             &82.3 &90.91  \\
\hline
BP data set &eig   &5         &   62.50             &79.5 &87.8  \\
\hline

\end{tabular}
\end{table*}

\subsection{Dataset}
The experiment is performed on 3 data sets out of which two data set are publicly available i.e. the Bearing dataset from the Prognostics Data Repository and second one is the MIMIC data-set for classifying blood pressure. The third data-set is internal to our organization and this aims to classify the emotion of a person into three classes namely happy, sad, and neutral. As there is the rest two data sets are two class classification problem, we have used our own data set to prove that our proposed system works on multi-class use-cases also. The first data set used as a machine automation case study. The rest two data sets are used in health care and emotion detection. 2nd and 3rd data sets are based on photoplethysmogram (PPG) sensor signal. The bearing data set (named IMS-Rexnord) consists of three datasets describing a test to failure experiment. Each dataset comprises of several files each of which has a record of 1 sec vibration signal snapshot which are recorded at specific intervals. The sampling rate is 20KHz, recording interval is 10 min and there are 20480 data points in each of the file. In our experiment we have used the second dataset. It has 984 files. Each individual file holds the record of 4 channels representing 4 bearings where the first bearing eventually turns faulty due to outer race failure.

The experiment is performed in two ways. In the first case only the bearing 1 data is considered. State of the art \cite{dong2014bearing} shows that the bearing 1 starts degrading after the 700th point where each file of 1 minute readings is denoted as a point. So here we have formulated a two class classification problem where the first 700 files are considered to be healthy and the rest 282 files are considered to be faulty (the last 2 files are discarded due to presence of many noisy signal values).

In the second case we have considered all the bearings to formulate the two class classification problem. The bearing dataset has 984 files. Each file has record of 4 bearings. Therefore the total number of good bearing samples is 3652 (1st 700 samples of 1st bearing and 984 samples of each of the three other bearings) and the number of faulty bearing samples is 282 (last 282 files of 1st bearing). To avoid any biasness we have segregated the data into 5 folds. Each of the dataset consisting of 282 faulty bearing samples and 730 good bearing samples.

The second dataset which is used to classify the blood pressure into high and low, records the PPG signal of 118 subjects from Bangalore and Gujarat. Among the 118 subjects, 15 subjects have high systolic blood pressure and 103 have low systolic blood pressure. To avoid any biasness due to imbalance in the dataset, the dataset is segregated into three datasets. Each of the dataset consists of the PPG signal record of 49 subjects, 15 subjects having high systolic blood pressure and 34 subjects having low systolic blood pressure. The sampling rate is 60 Hz.

The third dataset (used to classify the emotion into happy and sad) records the fingertip pulse oximeter data of 33 healthy subjects (13F and 20M) with average age 27. No two emotion elicitation video was shown to a subject in one single day. The Pulse Oximeter is used to detect and record the PPG signal. We used standard video stimuli which itself served as ground-truth and the rigorous experimentation procedure ensured that the time synchronization error between the stimuli and recorded physiological data is always less than 1sec.

Table 2 shows the performance of proposed method against state of the art results, proving the efficacy of the method. It is to be noted that time and cost is a huge factor when comparing the performance. In many cases, a trade-off fits the solution well when performance gain is not much when automotion is compared to the manual effort.

\subsection{Comparison against MLP}
Learning representations from raw data is an emerging field. Theano has been used as the software platform to carry out experiments.

Different number of layers (based on standard thumb rules of input size) for Multi-Layer Perceptron (MLP) with Dropout feature has been tried out so that automatic feature learning can take place.  Different activation functions like tan hyperbolic, softmax, sigmoid, rectified linear unit (relu) etc. has been investigated at different layer levels to get a suitable architecture for classification task for the given problems. 
 
Table 3 lists the configurations obtained for a 5 layered MLP for which the best performance was obtained. It is seen that MLP techniques fail in comparison to proposed method as well as state of the art. Also, MLP has some drawbacks when put into the IoT spectrum - a) need for a lot of data for training and b) non-interpretable feature extraction.

\subsection{Comparison against PCA}
Principal component analysis (PCA) is a statistical procedure that uses an orthogonal transformation to derive principal components representative of the features under consideration. This has two outcomes: a) dimension of feature space can be reduced by selecting most prominent principal components b) derived features is supposed to represent the feature space better.
Gaussian kernel is used for SVM based classification on the principal component features extracted. Various dimension reduction techniques has been used like Alternating Least Squares (als), Eigen Value Decomposition (eig) and the traditional Singular Value Decomposition (svd). Table 4 lists the configurations leading to the best results. It is apparent from the table that PCA based methods do not perform well in comparison to proposed method. Another drawback of PCA is the derived features are not interpretable.

\section{Conclusion}
This paper has carried out a survey to understand the workflow time and cost needs of a typical IoTA task.  In the survey, Feature Engineering came out as the most taxing task among all endeavors that comprise IotA. To tackle the challenge, a system was built to automate this sub-task of IoTA. The system has been tested on three datasets and has been found to give good results when compared to state of the art. The proposed method is compared with PCA and MLP, which are two divergent paths of feature engineering.  Feature interpretation is another notable aspect of the system. Future work will look into automation of parameter tuning and selection of machine learning models. Automatic window selection \cite{2013cikmSnehasis} for a given dataset is also planned. Integration of knowledge bases and use of reasoning \cite{2013patentDebnath} \cite{2013OTMdebnath} for ease of interpreting features and causality analysis is also planned.

\bibliographystyle{ACM-Reference-Format}

\balance

\end{document}